\documentclass{article}

\usepackage{microtype}
\usepackage[final]{graphicx}
\usepackage{booktabs} 
\usepackage{appendix}

\usepackage{float}
\usepackage[caption = false]{subfig}
\usepackage{adjustbox}
\usepackage{amsmath}

\usepackage[utf8]{inputenc}

\usepackage{hyperref}
\usepackage[noabbrev,capitalise]{cleveref}

\usepackage[accepted]{icml2020}

\icmltitlerunning{Automated Cleanup of the ImageNet Dataset}

\begin{document}

\twocolumn[
\icmltitle{Automated Cleanup of the ImageNet Dataset by \\ Model Consensus, Explainability and Confident Learning}



\begin{icmlauthorlist}
\icmlauthor{Csaba Kertész}{tpuni}
\end{icmlauthorlist}

\icmlaffiliation{tpuni}{TAUCHI Research Center, Faculty of Information Technology and Communication Sciences, Tampere University, Tampere, Finland}

\icmlcorrespondingauthor{Csaba Kertész}{csaba.kertesz@gmail.com}

\icmlkeywords{Machine Learning, ICML}

\vskip 0.3in
]



\printAffiliationsAndNotice{}  

\begin{abstract}
The convolutional neural networks (CNNs) trained on ILSVRC12 ImageNet were the backbone of various applications as a generic classifier, a feature extractor or a base model for transfer learning.
This paper describes automated heuristics based on model consensus, explainability and confident learning to correct labeling mistakes and remove ambiguous images from this dataset.
After making these changes on the training and validation sets, the ImageNet-Clean improves the model performance by 2-2.4 \% for SqueezeNet and EfficientNet-B0 models.
The results support the importance of larger image corpora and semi-supervised learning, but the original datasets must be fixed to avoid transmitting their mistakes and biases to the student learner.
Further contributions describe the training impacts of widescreen input resolutions in portrait and landscape orientations.
The trained models and scripts are published on Github (https://github.com/kecsap/imagenet-clean) to clean up ImageNet and ImageNetV2 datasets for reproducible research.
\end{abstract}

\section{Introduction}
\label{introduction}

The ILSVRC12 ImageNet dataset had an enormous impact on computer vision and its related fields in recent years. Convolutional neural networks built on ImageNet outperformed the hand-engineered feature descriptors \cite{bay2006surf, dalal2005histograms} and these pretrained CNNs were used for transfer learning. In this case, the intermediate latent features of the CNN are the input to train some additional layers for a certain new problem while the other layers are frozen. The latent features are powerful to encode natural images for various tasks not far from the ImageNet dataset distribution. Pretrained ImageNet models were successfully applied to robotic vision \cite{sunderhauf2018limits}, medical imaging \cite{raghu2019transfusion}, historical document analysis \cite{studer2019comprehensive}, classifying species \cite{van2018inaturalist}, cars \cite{krause20133d} or flowers \cite{nilsback2008automated}.

However, some challenges were uncovered for the ImageNet dataset with the image sampling, the labeling process and the learning methods. The dataset creation introduced several biases to the image distribution such as gender \cite{dulhanty2019auditing}, race \cite{stock2018convnets} and location \cite{shankar2017no}. The labeling process involved a human workforce to annotate the images that cannot guarantee the occasional errors. Regarding the learning methods, the models showed a lack of robustness to adversarial examples \cite{cisse2017houdini}.

Some attempts were made to correct the biases in human samples \cite{yang2020towards} and general labeling mistakes \cite{northcutt2019confident} in ImageNet, but their results were not analyzed for more CNNs and their proposed fixes were not published in a reusable form for other researchers.

This paper uses automated methods to clean up the ImageNet dataset and assess the effects of the modifications to deep learning training and validation performance.

\begin{figure*}
\centering
\subfloat[First][Label: polecat \\ \hspace*{0.14in} Real: mouse]{\label{fig1-a}\includegraphics[width = 1.3in]{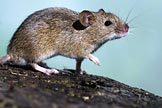}}\qquad
\subfloat[Second][Label: laptop \\ \hspace*{0.14in} Candidate: notebook]{\label{fig1-b}\includegraphics[width = 1.2in]{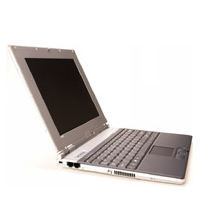}}\qquad
\subfloat[Third][Label: American\_egret]{\label{fig1-c}\includegraphics[width = 1.3in]{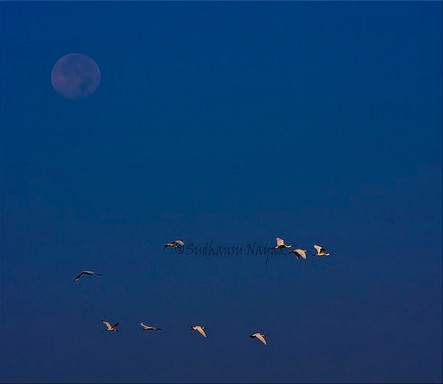}}\qquad
\subfloat[Fourth][Label: siamang]{\label{fig1-d}\includegraphics[width = 1.3in]{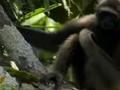}}\qquad
\subfloat[Fifth][Label: breakwater \\ \hspace*{0.13in} Candidate: seashore]{\label{fig1-e}\includegraphics[width = 1.4in]{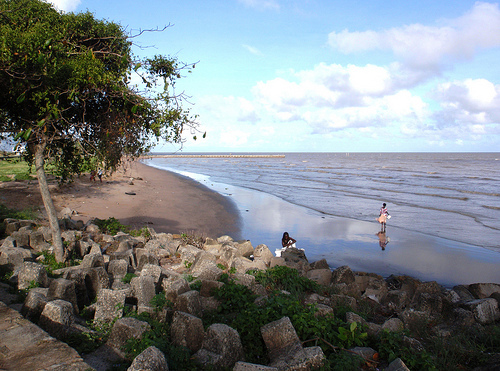}}\qquad
\subfloat[Sixth][Label: neck\_brace]{\label{fig1-f}\includegraphics[width = 1.3in]{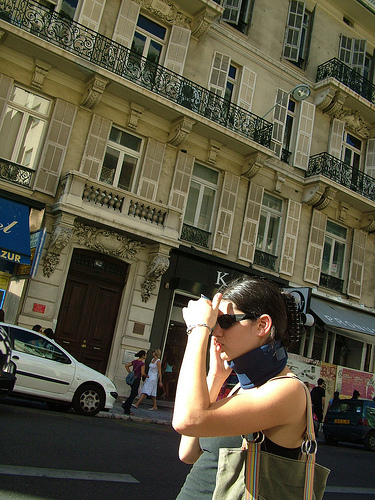}}\qquad
\subfloat[Seventh][Label: plate]{\label{fig1-g}\includegraphics[width = 1.3in]{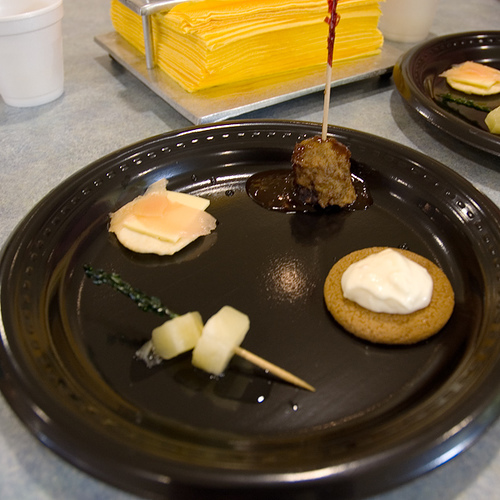}}\qquad
\subfloat[Eigth][Label: restaurant]{\label{fig1-h}\includegraphics[width = 1.2in]{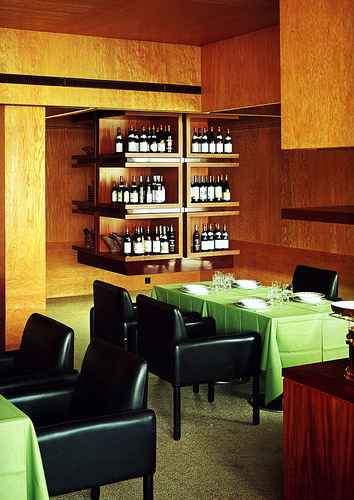}}
\caption{Labeling challenges in the ImageNet dataset: (a) incorrect label, (b) equivalent categories (c) miniscule objects, (d) bad quality photo, (e) multiple correct categories, (f) many objects, and (g-h) semantics.}
\label{fig1}
\end{figure*}

\section{Labeling Challenges in the ILSVRC12 ImageNet Dataset}
\label{dataset-challenges}

The ImageNet project annotated several million images into classes based on a synset hierarchy. The ILSVRC12 ImageNet corpus is a subset of this database and it contains over a million images for 1000 categories. One ground-truth class is assigned to each image in the database.
 
\cref{fig1} shows labeling problems from the ImageNet validation set. Although the labeling process evaluated every picture by more individuals on the Amazon Mechanical Turk service, wrong categories were sometimes assigned to the images. People did not recognize the content on the image, there were too many candidate categories present on the image or a similar, incorrect category was assigned. \cref{fig1-a} contains a vole and it was labeled as \emph{polecat}. The closest ILSVRC12 ImageNet category is \emph{hamster} while the closest ImageNet category is \emph{mouse}. Such images without matching ILSVRC12 ImageNet category should be removed from the dataset or changed to the closest category. \cref{fig1-b} displays a computer that is named as \emph{laptop} or \emph{notebook} in our everyday language. Nevertheless, these are different categories in the dataset what generates unnecessary noise for the training process. \cref{fig1-c,fig1-d} show two examples when the image content cannot be recognized by a human because either the subject is too small (\cref{fig1-c}) or it is too close to the camera with possible bad lightning conditions (Figure 1.d). Pictures with little information content can be removed from the dataset since they are not reasonable to be recognized without prior knowledge.

Even if a "correct" label is assigned, the image content can be ambiguous in many ways. \cref{fig1-e} shows when a scene has two equally important parts (a breakwater and a seashore). Both \emph{breakwater} and \emph{seashore} categories are correct and flipping the image label between these two categories are not wrong. However, too many categories are present on \cref{fig1-f} and the target object is small as well. The target label is \emph{neck\_brace}, but the image has additional objects for categories of \emph{pole}, \emph{terrace}, \emph{sunglasses}, \emph{car\_wheel}, \emph{car}, \emph{bracelet}. This diverse content can be confusing for trained models. In other cases, the whole scene is more important than a few smaller objects on the image. The dominant object on \cref{fig1-g} is a \emph{plate} though the correct category would change if the plate is almost fully covered by some food. Similarly, on \cref{fig1-h}, the interior of a \emph{restaurant} is more important meaning than labeling the image as \emph{desk}, \emph{plate} or \emph{bottle}.

These issues can be solved by fixing the labeling mistakes and assigning multiple classes to the images where it is applicable. Nonetheless, these corrections need tedious efforts and extensive funding to cover the costs. The author proposes an automated clean-up process by combining model consensus, confident learning and explainability to minimize human supervision. 

\section{Clean-Up Methods}
\label{clean-up methods}

The labeling challenges in ImageNet were identified by the research community \cite{beyer2020imagenet, northcutt2019confident, shankar2020evaluating} and initial attempts were made for mitigation. Northcutt et al \cite{northcutt2019confident} removed wrong samples detected by confident learning up to 10 \% of the training set and they could achieve 0.5 \% accuracy improvement with ResNet-50 on the ILSVRC2012 validation set. Beyer et al \cite{beyer2020imagenet} used pretrained models and crowdsourcing to relabel the ImageNet validation set (ReaL) with more possible classes per image. They removed appr. 10 \% of the training set, and after training for 90 epochs, the accuracy of ResNet50 improved by 0.3-0.4 \%. Some works developed new validation sets to test out-of-distribution samples. In \cite{hendrycks2020many}, artistic images were collected to form ImageNet-R validation set, while in \cite{recht2019imagenet}. new natural pictures were gathered by replicating the original labeling process to create ImageNetV2 validation sets. All these researches involved extensive human supervision in their processes except \cite{northcutt2019confident}. When images were removed by automated clean-up methods from the training set during the experiments in \cite{northcutt2019confident} and \cite{beyer2020imagenet}, the improvements were 0.3-0.4 \% for the original and ReaL validation sets. The author implemented a new process with label fixes and image removal. The former is beneficial to correct wrong labels without human intervention, the latter targets the deletion of noisy pictures for the training process. Namely, images with minuscule objects (e.g. \cref{fig1-c}), quality issues (e.g. \cref{fig1-d}) and too many categories (e.g. \cref{fig1-f}).

Confident learning, model consensus and explainability were utilized to get a good suboptimal solution whilst managing the time and computational constraints. An image sample is denoted by $s$ later in the paper, a set of $n$ images by $S_n = \{s_0, s_1, ..., s_{n-1}\}$, a label to a sample $s$ by $l$, a new label candidate to a sample $s$ by $l'$, the corresponding labels to $S$ by $L$, a deep learning model by $m$ and a set of $n$ models by $M_n = \{m_0, m_1, ..., m_{n-1}\}$. The predictions of a sample set $S$ by a model $m$ are denoted by $P_n = \{p_0, p_1, ..., p_{n-1}\}$. The clean-up process results in two sets. $S_r$ is a set of samples collected for removal from $S$ and $S_f$ is a set of samples whose annotation should be fixed by label candidates.

\subsection{Confident Learning}
\label{confident learning}

The confident learning theorem was invented by Northcutt et al in \cite{northcutt2019confident} to find label errors in datasets. It uses model predictions to estimate the uncertainty of the annotated labels and proposes new label candidates. If a sample $s$ is identified as mislabeled, a new label $l'$ is suggested instead of the original $l$. Confident learning is defined here as a $CL$ function:

\begin{equation}
\label{eq1}
\begin{gathered}
N'_m = CL(S, L, m, P, fn), \quad fn \in [0, 1] \\
(s'_i; l'_i) \in N'_m, \quad i \in \{0, ..., k-1\}
\end{gathered}
\end{equation}

where $CL$ takes a set of samples ($S$), annotated labels ($L$) and predictions ($P$) from model $m$ to output $k$ noisy samples with candidate labels ($|N'_m|=k$). The fraction noise ($fn$) is a hyperparameter between $[0, 1]$ to control the returned fraction of noisy labels per class. The fixed hyperparameters of confident learning were the pruning by noise rate and sorting the label indices by a normalized margin.

\begin{figure}[tp]
\centering
\subfloat{\includegraphics[width = 3.2in]{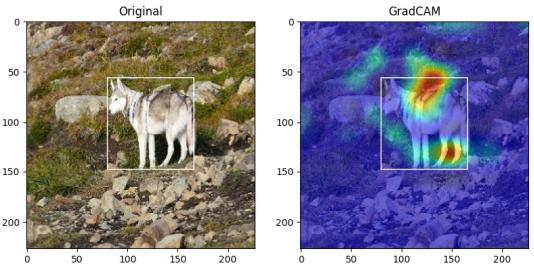}}
\caption{Explainability example. GradCAM heatmap visualizes the important regions for a CNN model to recognize the husky. The white rectangle is the annotated bounding box for the label. }
\label{fig2}
\end{figure}

For a set of models ($M$), the label candidates are gathered for each sample:

\begin{equation}
\label{eq2}
\begin{gathered}
(s_i, L'_i) \in N'_S, \quad \forall s \in S, \quad \exists (s, l') \in N'_m  \\
L'_i = (l'_0, l'_1, ..., l'_p), \quad 0 <= p < |M|
\end{gathered}
\end{equation}

where $N'_S$ contains the candidate label lists for each sample and the same label can occur multiple times in $L'_i$. The more times a sample is considered mislabeled by confident learning with the same candidate label the more likely the true label is this candidate. The annotation ($l_i \in L$) is fixed for a sample $s_i$ when the following conditions are met:

\begin{equation}
\label{eq3}
\begin{gathered}
|L'_i| >= h_1 \\
|unique\_items(L'_i)| < 3
\end{gathered}
\end{equation}

where $h_1$ number of models consider the sample mislabeled and there are less than three unique candidates. The sample $s_i$ and the most frequent candidate from $L'_i$ are added to $S_f$ in this case. A sample $s_i$ is added to $S_r$ for removal if the unique items in $L'_i$  are over a certain threshold $h_2$ ($|unique\_items(L'_i)| \geq h_2$). Samples marked for annotation fixes cannot be removed from $S$ ($(s_i, l'_i) \in S_f \Rightarrow s_i \notin S_r$)

To apply confident learning, the models $M$ and $fn$, $h_1$, $h_2$ must be defined. These latter hyperparameters can be found empirically after some initial experimentation with a dataset since confident learning is not a silver bullet. The author experienced that the success of confident learning depends on how representative is a training set and how good is a model built on it (see in \cref{validation-set}). In the upcoming \cref{cleaning-up-imagenet}, the official confident learning implementation (Cleanlab) was used from Github \footnote{URL: https://github.com/cgnorthcutt/cleanlab}.

\subsection{Top-5 Model Consensus and Explainability}
\label{model-consensus}

\cref{fig1} presents why a single label cannot describe some images in the ImageNet dataset unequivocally when the top-1 accuracy is calculated. Nevertheless, the top-5 accuracy lets a model assign five potential labels for an input picture which can recognize a single image with a few categories. On the other hand, if an image is not predicted by the top-5 predictions correctly, that picture likely exhibits one challenge from \cref{fig1-c,fig1-d,fig1-f} (minuscule subject, bad image quality, more than five categories). These images would be beneficial to remove thus a process is proposed to find these pictures indirectly by the top-5 mispredictions and explainability.

Fort et al \cite{fort2020deep} showed that the deep learning models do not learn the same function after their training runs. They produce similar accuracy with the same dataset and hyperparameters, but they disagree about the actual misclassified samples. In other words, when models built on a training set achieve appr. 90 \% accuracy, each model misclassifies different samples in the validation set (10 \% error). Because of the misclassified samples change for the same model architecture after every training run, various topologies and model capacities must behave likewise. If most models in an $M$ miss certain images to classify correctly in their top-5 predictions, we can assume that those images are good candidates for removal. Let us denote a counter variable ($C_s$) for every sample ($s$) to sum all models which do not predict the label ($l$) in their top-5 predictions:

\begin{equation}
\label{eq4}
\begin{gathered}
 \\
c_s(m, s, l) = \begin{cases} 1 & \mbox{if } l \notin top_5(m(s)) \\0 & \mbox{otherwise} \end{cases} \\
C_s =  \sum^{|M|-1}_{i=0} c_s(m_i, s, l) \\
\end{gathered}
\end{equation}

Those images will be added to $S_r$ whose $C_s \geq h_3$. However, such an images can be exempted from removal if at least one model detects the target object with reasonable confidence by explainability.

There are various interpretability algorithms, but the author found only some class activation mappings (CAMs) reliable and quick enough for large-scale batch processing. GradCAM \cite{selvaraju2017grad}, GradCAM++ \cite{chattopadhay2018grad} and ScoreCAM \cite{wang2019score} were implemented based on a GitHub project \footnote{URL: https://github.com/tabayashi0117/Score-CAM}. These algorithms provide a heatmap to show which pixels contribute to the actual decision of a neural network (\cref{fig2}). The hottest areas (red pixels close to value 1) have a major contribution to the model output while the neutral pixels are close to zero values (blue areas). The criteria for an image to avoid the removal by insufficient top-5 predictions if at least 1 \% of the bounding box area of the target subject (white rectangle on \cref{fig2}) have 0.75 values on the heatmap when a good model prediction is analyzed. Visually, this condition can be translated so that at least 1 \% of the bounding box area is red. The \emph{explainability score} will mean the percentage of pixels in a bounding box whose value is at least 0.75 on the heatmap. The husky on \cref{fig2} has an explainability score 0.06 (6 \%) by GradCAM. Finally, three CAMs make the final decision on the images because these methods are not reliable enough alone, they produce varying results (see \cref{app-fig1} in Appendix), similar to the experiences in \cite{graziani2021evaluation}. Thus, at least two methods out of the three CAMs must met the 1 \%-red-area condition for the bounding box for a model to exempt the inclusion in $S_r$ if $C_s \geq h_3$. Since the explainability methods visualize a model decision, those images can be examined that have at least one successful model prediction in their top-5 results. If the bounding box is not available for a sample, the explainability step is omitted in the process.

\subsection{Merged Categories}
\label{merged-categories}

Some categories in ILSVRC12 ImageNet are interchangeable. Either two categories are synonyms or the difference in their meaning is so subtle that the images were highly intermixed during the labeling process. More studies identified problematic categories \cite{beyer2020imagenet, northcutt2019confident, tsipras2020imagenet} in the past, but a comprehensive solution was not applied to check the effects on the training and the model performance. The author gathered these past experiences and complemented them with new categories found during the initial experimentation (see \cref{app-table1} in Appendix). The categories were fixed by copying the samples of one category into a close/synonym category. After moving the samples, a placeholder black image was generated into the empty category to have one placeholder sample. This latter step kept the original 1000-class interface of the ImageNet models compatible with the newly-trained models after the clean-up process. The category merging does not have a foreseeable larger impact on the model performance since the affected samples are appr. 0.04 \% of the dataset, but the training is benefited by fewer overlapping categories and eliminated interclass variability between these confusing categories. The categories were merged after the image fixes and removal steps in the clean-up process.

\section{Cleaning Up ImageNet datasets}
\label{cleaning-up-imagenet}

ILSVRC2012 ImageNet dataset has two parts, the training set ($T$) contains 1281167 images and 523966 of them have bounding boxes specified for the image label. The original validation set ($V$) contains 50000 images and all of them have bounding box annotations. The ImageNetV2 validation sets were developed by Recht et al \cite{recht2019imagenet}, the matched frequency version ($V2$) was included in the experiments in this section. It contains 10000 images, 10 per category. They were collected by replicating the same labeling procedure of $T$ and $V$, but there are no bounding box annotations for these images. The datasets were cleaned up separately, the cleaned training set will be denoted by $T_c$, the cleaned original validation set by $V_c$ and the cleaned ImageNetV2 with matched frequency by $V2_c$.

\subsection{ILSVRC12 Validation Set}
\label{validation-set}

The author chose 10 pretrained ImageNet models to build a consensus on which images should be fixed or removed. The model set $M_{val}$ had lightweight CNN models from recent years: SqueezeNet, NASNet, MobileNet, MobilenetV2 and EfficientNet (B0, B1, B2 and their Noisy Student variants). These pretrained models were fetched from Keras (Tensorflow) except SqueezeNet \footnote{URL: https://github.com/rcmalli/keras-squeezenet} and EfficientNet \footnote{URL: https://github.com/qubvel/efficientnet}. There were three reasons why bigger models were not included in this set. First, explainability methods were executed on many images, and practically, it is not feasible to run them with bigger models on this scale. Secondly, the lighter models generalize better on mislabeled images as it was found in \cite{northcutt2020labelerrors} recently. Finally, the author had the intuition that the latest semi-supervised models reached the limits of top-1 accuracy on the ImageNet dataset because of its labeling problems described in \cref{dataset-challenges}. These big models might not really learn better abstractions, but guessing better by other shortcuts (e.g. using the background to bet on the foreground object). Beyer et al \cite{beyer2020imagenet} suggested in their analysis of co-occurring labels that the noisy student versions of EfficientNet exploit the biases in the original ImageNet labeling procedure.

\begin{table}[tp]
\caption{Label errors found by confident learning per class ($N'_m$) as well as pretrained model accuracies on the original ImageNet validation set ($V$) and the cleaned version ($V_c$). The last four rows have results for some large models that were not used in the clean-up process hence their label errors are empty.}
\label{table1}
\vskip 0.1in
\begin{center}
\begin{adjustbox}{width=\columnwidth,center}
\small
\begin{sc}
\begin{tabular}{lcccr}
\toprule
Model                & Label  & Top-1/Top-5 & Top-1/Top-5 \\
                     & errors & Original    & Cleaned    \\
\midrule
SqueezeNet           & 13716 & 56.15/79.38 & 63.78/85.66 \\
MobileNet            & 8550 & 68.88/88.60 & 78.14/94.45 \\
MobileNet V2         & 7341 & 69.70/89.35 & 79.26/95.09 \\
NASNet               & 7489 & 71.83/90.49 & 81.64/95.88 \\
ENet-B0              & 5422 & 76.22/92.96 & 86.61/98.00 \\
ENet-B0 (NS)         & 5597 & 76.82/93.65 & 87.09/98.40 \\
ENet-B1              & 5224 & 77.81/93.86 & 87.98/98.53 \\
ENet-B1 (NS)         & 4926 & 80.20/95.33 & 89.91/99.15 \\
ENet-B2              & 4921 & 79.06/94.54 & 89.26/98.89 \\
ENet-B2 (NS)         & 4658 & 81.41/95.94 & 90.73/99.37 \\
\midrule
ENet-B5 (NS)         & -    & 86.07/97.75 & 90.23/99.19 \\
ENet-B7              & -    & 84.93/97.20 & 90.11/99.10 \\
IG-ResNext101        & -    & 85.42/97.88 & 89.06/98.86 \\
ENet-L2 (NS)         & -    & 88.35/98.65 & 90.87/99.25 \\
\bottomrule
\end{tabular}
\end{sc}
\end{adjustbox}
\end{center}
\vskip -0.2in
\end{table}

Since these models in $M_{val}$ were pretrained on the whole ImageNet dataset and all images had bounding box annotations, the hyperparameters of the clean-up process were less strict. Half of the models were sufficient to fix a label by confident learning ($h_1 = |M_{val}| / 2$), all candidate labels were considered by $fn = 1$ and the removal was initiated over 2 label candidates ($h_2 = 3$). The top-5 model consensus hyperparameter for removal was $h_3 = |M_{val}|*2 / 3$.

The models made no mistake for 74.45 \% of $V$ regarding their top-5 predictions. This top-5 consensus error on the validation set (25.54 \%) was similar to the problematic ratio (29 \%) of multi-object and multicategorical images in \cite{beyer2020imagenet}. Eventually, 3565 labels were fixed by confident learning and 2471 images were removed. These changes involved 7.13 \% and 4.9 \% of the original validation set respectively.

\cref{table1} shows the label errors found by confident learning and the clean-up results with the pretrained models. Most label errors were found for SqueezeNet, the least for EfficientNet-B2 and a strong correlation can be identified between the decreasing error count and the growing model complexity. The better a model learns the abstractions in the training set the fewer label errors are detected by confident learning. Similar to these findings, Shaikhina et al \cite{shaikhina2021effects} found across various machine learning classifiers and datasets that the post-hoc techniques have degraded quality when the models are uncertain in their decisions. However, it is a good practice to avoid relying on the larger models exclusively since the models learn different prediction functions \cite{fort2020deep} and the smaller models generalize better on unseen mislabeled samples \cite{northcutt2020labelerrors}. The middle column of \cref{table1} contains the top-1/top-5 accuracies on the original $V$ and the top-1 results improved by 7-11 \% while the top-5 had a 4-6 \% increase. The 99 \% top-5 accuracies on $V_c$ suggest upper adaptability limits that are already reached by the smaller EfficientNet models for the original ImageNet validation set. Since multicategorical images are still present after the clean-up process, it cannot be expected to reach 100 \% top-1 accuracy. The author evaluated four large models from \cite{rw2019timm}, close to the latest state-of-the-art (SOTA) results on ImageNet (last rows of \cref{table1}). Their top-1 accuracies did not go above the EfficientNet-B2 performance (appr. 90 \% accuracy). Some examples label fixes and removed images are shown in the Appendix (\cref{app-fig2}).

\subsection{ILSVRC12 Training Set}
\label{training-set}

The training set had to be cleaned by another approach compared to the validation set because the confident learning requires unseen samples for the label error estimation. The training set was evaluated by 4-fold cross-validation in Cleanlab to get label corrections for the whole set based on the testing folds. When the cross-validation was run on $T$, three folds were used for training which is 25 \% fewer training images compared to the whole $T$. The decreased training set size impacts the model performance negatively during the cross-validation as Hernández-García and König shown in \cite{hernandez2018data}, the model accuracy can drop up to 12-15 \% after removing half of the ImageNet training set. The smaller model accuracy means more uncertain estimation of the label errors on the testing folds insomuch as this relationship was discovered on \cref{table1}. This effect can be minimized by increasing the fold count, but that solution inflates the computational complexity dramatically since the ImageNet training is resource demanding.

8 deep learning models ($M_{tr}$) were crossvalidated: SqueezeNet, NASNet, MobileNet, MobileNetV2, ResNet18, EfficientNet-B0, EfficientNet-B1 and EfficientNet-B2. Because of the decreased model confidence, stricter hyperparameters were defined. The fraction noise remained $fn = 1$, but all models had to agree in a label fix ($h_1 = |M_{tr}|$) and the removal was over more label candidates ($h_2 = |M_{tr}| / 2$). The top-5 model consensus hyperparameter for removal was tightened to $h_3 = |M_{tr}|$. The stricter hyperparameters for removal were needed because the initial experimentation shown that removing 8-15 \% problematic samples result in 1.5-2.5 \% performance loss. Northcutt et al reported a similar loss for ResNet18 after removing 15 \% of of the training set in \cite{northcutt2019confident}. The source of this phenomena is that even the problematic images contribute to feature learning despite their noisy labels. Finally, 110839 labels were fixed by confident learning and 48581 images were removed from the training set. These numbers correspond to 8.6 \% and 3.7 \% of the ImageNet training set respectively.

\subsection{ImageNetV2 Matched Frequency Set}
\label{imagenetv2-set}

The ImageNetV2 matched frequency set ($V2$) have 10000 new images (10 images/category) sampled 10 years after the original ImageNet release \cite{recht2019imagenet}. Since the sampling process was replicated to collect the images under similar conditions, the same limitations and labeling errors can be expected from $V2$. The state-of-the-art ImageNet models achieve a maximum of 80.50 \% top-1 accuracy on $V2$, therefore, the ImageNet models were less capable to estimate the labeling errors and the removable images compared to $V$. ImageNetV2 lacks annotated bounding boxes what prevents the use of interpretability for the top-5 model consensus. The pretrained models of $M_{val}$ were utilized to clean up this validation set. Considering these challenging conditions, the most strict hyperparameters were defined in the clean-up process. The fraction noise was decreased to $fn = 0.9$, but other parameters for confident learning and top-5 model consensus remained the same ($h_1 = |M_{val}|$, $h_2 = |M_{val}| / 2$, $h_3 = |M_{val}|$). These hyperparameters resulted in 292 label fixes and 524 image removals that are 2.9 \% and 5.2 \% of ImageNetV2 respectively. Some example label fixes and removed images are shown in the Appendix (\cref{app-fig3}) to demonstrate the repeated labeling mistakes in ImageNetV2 and the original ImageNet validation set.

\begin{table*}[th]
\caption{The effects of cleaning up the training and validation sets for ImageNet. The first three rows contain the results for SqueezeNet 1.1, ShuffleNetV2 x1.5 and EffecientNet-B0 models trained and evaluated in different settings. The column headers show the used training/validation sets and the numbers are top-1/top-5 accuracies. The last four rows show pretrained models built on the original ImageNet training set and evaluated on the clean validation sets. Therefore, the last two columns are special cases where the models in the upper three rows were trained on the cleaned training set while the other rows were trained on the original training set. ($T$ - ImageNet training set, $T_c$ - cleaned $T$, $V$ - ImageNet validation set, $V_c$ - cleaned $V$, $V2$ - ImageNetV2 matched frequency validation set, $V2_c$ - cleaned $V2$)}
\label{table2}
\vskip 0.10in
\begin{center}
\begin{adjustbox}{width=\columnwidth+2.3in,center}
\small
\begin{sc}
\begin{tabular}{lcccccr}
\toprule
Model                & Top-1/Top-5 & Top-1/Top-5 & Top-1/Top-5 & Top-1/Top-5 & Top-1/Top-5  & \\
                     & $T, V$      & $T,V_c$     & $T_c,V_c $  & $T_c/T,V2$  & $T_c/T,V2_c$ & \\
\midrule
SqueezeNet 1.1       & 55.07/78.19 & 60.99/83.98 & 63.15/84.71 & 42.69/65.15 & 47.72/71.38  & \\
ShuffleNetV2         & 70.39/89.25 & 77.94/94.56 & 78.31/94.50 & 56.64/77.81 & 63.07/84.32  & \\
ENet-B0              & 73.06/91.40 & 80.74/96.24 & 83.11/96.78 & 61.53/82.23 & 68.58/88.87  & \\
\midrule
ENet-B5 (NS)         & 86.07/97.75 & 90.23/99.19 & -           & 76.70/93.51 & 81.73/96.58  & \\
ENet-B7              & 84.93/97.20 & 90.11/99.10 & -           & 74.45/92.12 & 79.49/95.73  & \\
IG-ResNext101        & 85.42/97.88 & 89.06/98.86 & -           & 76.77/93.42 & 80.68/96.04  & \\
ENet-L2 (NS)         & 88.35/98.65 & 90.87/99.25 & -           & 80.10/95.86 & 84.12/97.81  & \\
\bottomrule
\end{tabular}
\end{sc}
\end{adjustbox}
\end{center}
\vskip -0.2in
\end{table*}

\begin{table}[th]
\caption{Testing SqueezeNet and EfficientNet-B0 models with different input resolutions on the cleaned ImageNet validation set ($V_c$). The first two resolutions are in landscape orientation and the rest are in portrait.}
\label{table3}
\vskip 0.10in
\begin{center}
\begin{adjustbox}{width=\columnwidth-0.3in,center}
\small
\begin{sc}
\begin{tabular}{lcccr}
\toprule
Resolution    & Top-1/Top-5 & Top-1/Top-5 & \\
              & SqueezeNet  & ENet-B0     & \\
\midrule
320x180 (L)   & \textcolor{red}{-0.56}/\textcolor{red}{-0.05} & \textcolor{blue}{+0.26}/\textcolor{blue}{+0.09} & \\
384x216 (L)   & \textcolor{blue}{+0.93}/\textcolor{blue}{+0.89} & \textcolor{blue}{+1.36}/\textcolor{blue}{+0.75} & \\
180x320 (P)   & \textcolor{red}{-1.03}/\textcolor{red}{-0.30} & \textcolor{blue}{+0.25}/\textcolor{blue}{+0.33} & \\
216x384 (P)   & \textcolor{blue}{+0.29}/\textcolor{blue}{+0.48} & \textcolor{blue}{+1.30}/\textcolor{blue}{+0.55} & \\
\bottomrule
\end{tabular}
\end{sc}
\end{adjustbox}
\end{center}
\vskip -0.2in
\end{table}

\section{New Models}
\label{new-models}

After the clean-up process was finished, new models were trained to examine the effects on the model performance. All training sessions had the same hyperparameters to avoid any influence by such changes when the results were compared to each other. The training was executed by pytorch-image-models \cite{rw2019timm} for 90 epochs with cosine scheduler, one warm-up epoch, 10 cool-down epochs, initial learning rate 0.1, random pixel erase probability 0.5 \cite{zhong2018randomerase}, SGD loss function and RandAugment \cite{cubuk2019randaugment} data augmentation (magnitude 7, the standard deviation of magnitude noise 0.5, increased severity with magnitude). These hyperparameters were not the most optimal to get accuracy close to the SOTA for any models, but they provided reasonably good results to compare the effects of the clean-up process. The default input resolutions were the standard square 224x224.

Three deep learning architectures were explored during the experiments, SqueezeNet 1.1, ShuffleNetV2 x1.5 and EfficientNet-B0. These models are small enough to train in many configurations, nevertheless, they scale in complexity and accuracy from SqueezeNet up to EfficientNet-B0. \cref{table2} shows the model performance for different training and validation set configurations. After cleaning up the ImageNet validation set, the third column contains the performance gains. The small models achieved +5-7.5 \% top-1/top-5 accuracies compared to $V$ while the big pretrained models hit accuracy limits, 90 \% for top-1 and 99 \% for top-5 after minimal gains. When the small models were trained on a cleaned training set ($T_c$), further +0.3-2.3 \% accuracies were reached in the fourth column. When Beyer et al \cite{beyer2020imagenet} tested a ResNet50 classifier built on their cleaned ImageNet training set, they experienced +0.3 \% accuracy improvement against their cleaned validation set (ImageNet-ReaL). SqueezeNet 1.1 and EfficientNet-B0 achieved a bigger, +2.16 \% and +2.37 \% gains respectively. The sixth and seventh columns show the impact of the cleaned ImageNetV2 validation set ($V2_c$). The small models got improved top-1 accuracies by +5-7 \% and better top-5 accuracies by +6-6.5 \%. The big pretrained models built on original $T$ had lower gains, +4-5 \% top-1 and +2-3 \% top-5 accuracies.

Some widescreen resolutions were investigated during the experiments since ImageNet models are always trained in square input shape, but the webcams and phone cameras have a widescreen aspect ratio. The results for two portrait and landscape resolutions are shown on \cref{table3} after SqueezeNet 1.1 and EfficientNet-B0 models were trained on $T_c$, tested on $V_c$. The smaller resolutions (320x180, 180x320) have approximately the same amount of pixels as the standard input resolution (224x224). The bigger resolutions have 1.65x more pixels than the standard input. The EfficientNet-B0 models had a positive gain in every configuration, appr. +0.25 \% top-1 accuracy for the smaller resolutions and +1.3 \%  for the bigger. For SqueezeNet model, the smaller resolutions caused performance degradation while the bigger resulted in +0.2-0.9 \% gains. The increased pixel count always improved the model performance, but the resolutions 320x180/180x320 caused problems for the SqueezeNet model. All negative accuracy changes originated in the different cropping mechanisms during the training and evaluation compared to the usual square cropping. Nevertheless, the availability of widescreen models is an obvious advantage to use the camera pictures without resizing or cropping to square aspect ratio in real-world applications. The other takeaway of \cref{table3} is the accuracy increase by raising the input resolution for the existing architectures. In this case, the internal model parameter count does not change and the model will have minimal runtime cost. The bigger widescreen resolution changed the inference times by +10-20 msec on an Intel i7-7500U CPU (SqueezeNet: 22 msec$\rightarrow$34 msec, EfficientNet-B0: 100 msec$\rightarrow$121 msec) and the difference could not be measured with GPU inference.

\begin{table}[tp]
\caption{Testing models on ImageNet-R and cleaned ImageNetV2 validation sets after adding cleaned EDSR and CAE samples to the training set.}
\label{table4}
\vskip 0.10in
\begin{center}
\begin{adjustbox}{width=\columnwidth-0.2in,center}
\small
\begin{sc}
\begin{tabular}{lcccr}
\toprule
Model                & Top-1/Top-5 & Top-1/Top-5 & \\
                     & ImageNetV2  & ImageNet-R  & \\
\midrule
SqueezeNet 1.1       & \textcolor{blue}{+0.76}/\textcolor{blue}{+0.47} & \textcolor{blue}{+5.39}/\textcolor{blue}{+6.90} & \\
ShuffleNetV2         & \textcolor{red}{-0.65}/\textcolor{red}{-0.66} & \textcolor{blue}{+5.96}/\textcolor{blue}{+6.33} & \\
ENet-B0              & \textcolor{red}{-1.49}/\textcolor{red}{-1.02} & \textcolor{blue}{+5.86}/\textcolor{blue}{+5.73} & \\
\bottomrule
\end{tabular}
\end{sc}
\end{adjustbox}
\end{center}
\vskip -0.2in
\end{table}

Hendrycks et al \cite{hendrycks2020many} introduced a new validation set (ImageNet-R) to test out-of-sample distribution and they published modified ImageNet training sets augmented by their DeepAugment method. They distorted $T$ by two models (EDSR \cite{lim2017enhanced} and CAE \cite{theis2017lossy}) which resulted three times more training data in the end. \cref{table4} shows the accuracies when the models were trained on this big training set and validated on ImageNet-R and cleaned ImageNetV2 ($V2_c$). Surprisingly, the increased training set improves the performance by appr. +6 \% accuracy on ImageNet-R for all models, but it was counterproductive on $V2_c$ in the function of the model complexity. The worse the performance on ImageNetV2 the bigger the model. The reason is that ImageNet-R contains \emph{artistic} out-of-distribution images (sketches, cartoons, paint etc.) while ImageNetV2 has \emph{natural} out-of-distribution images. CAE and EDSR modified the original ImageNet training set ($T$) thus the new images were relatively close to the originals in the feature subspace. They can be considered like local perturbations of the original images. As \cite{fort2020deep} showed the near solutions around a local, suboptimal minimum on the loss landscape are functionally close to each other. DeepAugment helps to learn the \emph{artistic representations} of $T$, but it does not improve the generalization to natural out-of-distribution images. It facilitates a better local solution around a minimum on the loss landscape by modeling the local perturbations similar to Bayesian networks instead of increasing the generalization power in the global feature space \cite{fort2020deep}.

\section{Discussion}
\label{discussion}

Northcutt et al \cite{northcutt2020labelerrors} examined ten machine learning datasets from different domains (computer vision, natural language processing) to discover the labeling errors. They found smaller capacity models performed worse on the test sets because of the lack of their learning skills while they generalized better on mislabeled samples in the validation set. Models with more capacity behaved in the opposite way. They proposed that larger models might overfit the noise in the training data in order to explain their findings and they suggested the performance on mislabeled samples as a measure for overfitting. However, the author believes their results were caused by a more straightforward reason. The sample collection process for the training and validation sets in a certain dataset is identical, therefore, similar labeling errors will be present in both sets (see \cref{app-fig2} and \cref{app-fig3} in the Appendix). The lower capacity models generalize on mislabeled, unseen samples better because they cannot learn the fine-grained patterns in the training set. The larger models do not overfit to the mislabeled samples, but they \emph{learn} them. In \cite{graziani2019interpreting}, the deep learning models learned the real data distributions in the earlier epochs during the training process while the mislabeled samples were fit in later stages. These findings make sense because the mislabeled samples act as outliers for a category in the feature space and a larger model learns the wrong samples better with its bigger model capacity.

Cleaning up the training datasets is a necessary thing for the larger models to learn good generalization power instead of the labeling errors and clean validation sets can verify this power. However, the hidden benefits of mislabeled samples cannot be overlooked. Although these samples have wrong labels, they contribute to the feature learning in the deep learning models. When 10 \% of labeling error candidates were removed from the ImageNet training set in \cite{northcutt2019confident}, a degradation in the validation performance was observed. The author found similar effects during the initial experimentation when 5-15 \% wrong sample candidates were removed from the training set, the result was 0.5-1.5 \% accuracy loss on the validation set. These results suggested for the author that the labeling fixes and the sample removal are necessary for a better dataset, but these methods must be balanced. Too many fixes can change false positive candidates to a wrong label more likely while too many samples cannot be removed from the training set to avoid negative effects of losing generalization power.

Using the top-1 accuracy to measure the model performance on ImageNet is challenging. The ImageNet dataset assigns one category to each image and the top-1 accuracy tests the model prediction against this label. Although the top-1 and top-5 accuracies are reported for the ImageNet models, the top-1 accuracy is the common measure to rank the models and determine the state-of-the-art. Beyer et al \cite{beyer2020imagenet} found 29 \% of their examined ImageNet samples contained multiple objects or they could be assigned to synonym categories. Because of so many ImageNet images have ambiguous labels, the top-1 accuracy cannot be a true measure for the models. If the mislabeled images are additionally considered in this equation, it is questionable how much the top-1 accuracies over 80 \% accuracy can mean a valid rank for the models. The author agrees with the opinion in \cite{stock2018convnets}, the top-5 accuracy is a more valid measure for the models until ImageNet is properly fixed and gets multilabel annotations.

Since the original training set in ImageNet has many mislabeled samples, the teacher transfers this knowledge to the learners in semi-supervised training. The validation set has similar mislabeled images which implicitly rewards the students to learn these inaccuracies from the training set. A cleaned training set benefits the semi-supervised learning by teaching better generalization on an unlabeled dataset and a cleaned validation set can verity the potential gains from the massive training dataset. Currently, the big model capacities in semi-supervised learning might be used to learn the local mislabeling strategies of ImageNet in the feature space instead of helping the generalization. An iterated clean-up process on the ImageNet dataset would benefit better semi-supervised models for the future.

\subsection{Limitations}
\label{limitations}

There are a number of limitations in this research. First of all, the clean-up process was executed once, but this automated procedure should be repeated more times to improve the training and validation sets incrementally. The more the trained models are built and evaluated on clean sets the more improvements can be gained in feature learning. Another improvement would be the human verification before removing the images since the current approach eliminates problematic images, but these steps can remove out-of-sample images as well.

\section{Acknowledgements}
\label{acknowledgements}

The author thanks the Finnish supercomputing provider (CSC - IT Center for Science) to give access to on-demand GPU and CPU resources for this research.


\bibliography{icml_paper}
\bibliographystyle{icml2020}


\onecolumn
\appendix

\begin{center}{\Huge Appendix}\end{center}
\setcounter{figure}{0}
\renewcommand{\thefigure}{A.\arabic{figure}}
\setcounter{table}{0}
\renewcommand{\thetable}{A.\arabic{table}}

\section{Explainability Examples from the ILSVRC12 ImageNet Validation Set}

\begin{figure*}[htp]
\centering
\subfloat[First][Label: maillot, model: SqueezeNet, explainability scores = (0.2, 0.1, 0)  \\ \hspace*{0.16in}Top-5 predictions: punching bag, bathtub, bikini, tub, maillot]{\label{app-fig1a}\includegraphics[width = 6.3in]{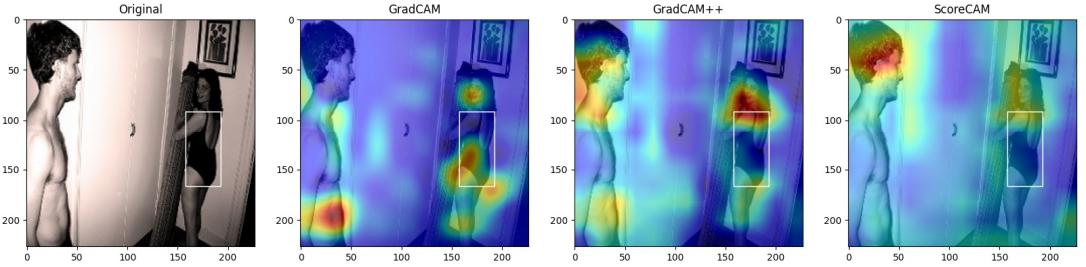}} \\

\subfloat[Second][Label: water jug, model: SqueezeNet, explainability scores = (0, 0.05, 0.07)  \\ \hspace*{0.16in}Top-5 predictions: milk can, caldron, rain barrel, pot, water jug]{\label{app-fig1b}\includegraphics[width = 6.3in]{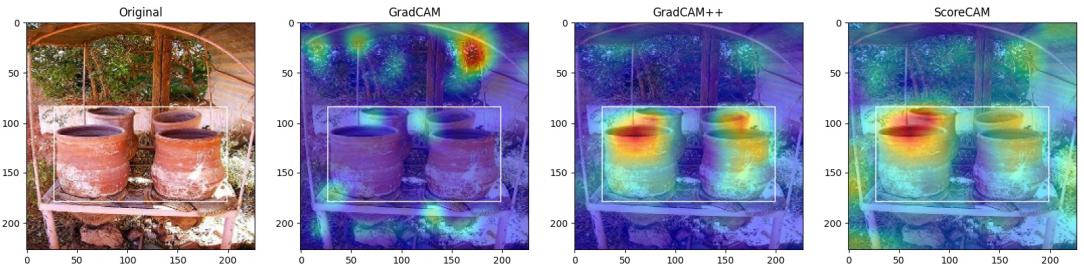}} \\

\subfloat[Third][Label: cassette, model: Efficient-B1 (Noisy Student), explainability scores = (0, 0, 0)  \\ \hspace*{0.16in}Top-5 predictions: tape player, cassette player, radio, cassette, loudspeaker]{\label{app-fig1c}\includegraphics[width = 6.3in]{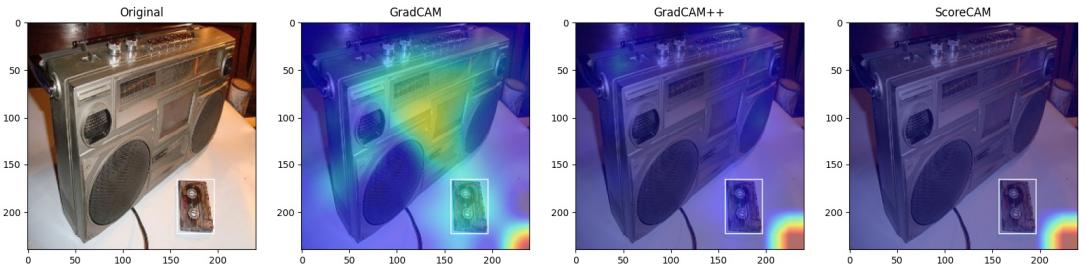}} \\

\caption{Three explainability methods (GradCAM, GradCAM+, ScoreCAM) are shown by heatmaps for images from the ILSVRC12 validation set. White rectangles show the annotated bounding boxes of the labels. The label, deep learning model, explainability scores and top-5 predictions are listed for every image. The figure demonstrates the agreement/disagreement inbetween the explainability methods: a) all methods disagree, b) only GradCAM++ and ScoreCAM agree, c) all methods fail for EfficientNet-B1 (Noisy Student). }
\label{app-fig1}
\end{figure*}

Graziani et al \cite{graziani2021evaluation} showed that CAMs cannot be applied to decision making in histopathology because they disagree on their heatmaps about the dominant image regions in classifier predictions. \cref{app-fig1} presents how unreliable are the explainability methods for ImageNet sometimes under the same conditions. The author used three methods: GradCAM, GradCAM++ and ScoreCAM. Usually, at least two of these methods work and they show where a classifier concentrates on an image to make the prediction. However, these class activation maps (CAMs) are not always uncover the same conclusions. \cref{app-fig1a} is a case when SqueezeNet gives the top-5th prediction as \emph{maillot} correctly, but all CAMs suggest a different region on the image as a basis for the decision. Since GradCAM++ is an evolved version of GradCAM, they share common roots and they give similar results many times, but \cref{app-fig1b} shows an image when GradCAM++ and ScoreCAM give close results while the red regions on GradCAM's heatmap are off from the bounding box of the target subjects. The author noticed a phenomenon with CAMs specific to deep learning models trained with the Noisy Student (NS) method. Namely, CAMs tend to suggest for these models that the prediction depends on one of the image corners. On \cref{app-fig1c}, all CAMs visualize that EfficientNet-B1 (NS) bases its decision on the bottom-right image corner. It is an open question if CAMs fail with models trained by semi-supervised learning or if they show the real decision what causes these NS models to decide based on the image corner?

Another takeaway of the \cref{app-fig1} results that the classifiers group similar categories together when they learn. SqueezeNet predictions were punching bag, bathtub, bikini, tub, maillot in \cref{app-fig1a} and the last four are closely related to pictures around swimming pools. The punching bag top-1 prediction makes sense because a half-naked guy is on the left side and he is typical object around punching bags in a gym. This scenario shows that the deep learning models guess the labels from the surroundings instead of the actual object to be detected. In \cref{app-fig1b}, SqueezeNet predicts milk can, caldron, rain barrel, pot and water jug. These objects are all similar to a water jug pottery in the feature space of how they look like. Finally, EfficientNet-B1 predicts categories close to older radios, cassette players: tape player, cassette player, radio, cassette, loudspeaker.

\section{ILSVRC12 ImageNet Validation Set Fixes}

\begin{figure*}[ph]
\centering

\subfloat[First][Old label: sports car \\ \hspace*{0.14in} New label: racer car]{\label{app-fig2a}\includegraphics[width = 1.3in]{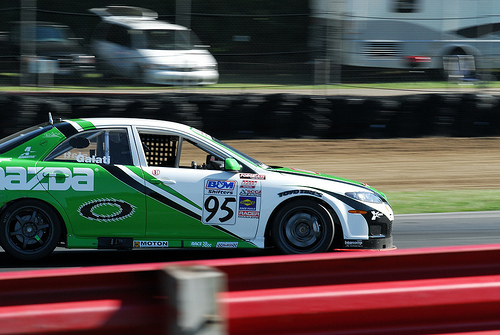}}\qquad
\subfloat[Second][Old label: table \\ \hspace*{0.14in} New label: restaurant]{\label{app-fig2b}\includegraphics[width = 1.3in]{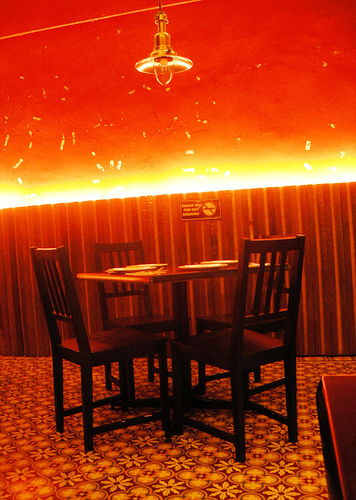}}\qquad
\subfloat[Third][Old label: mouse \\ \hspace*{0.14in} New label: mousetrap]{\label{app-fig2c}\includegraphics[width = 1.3in]{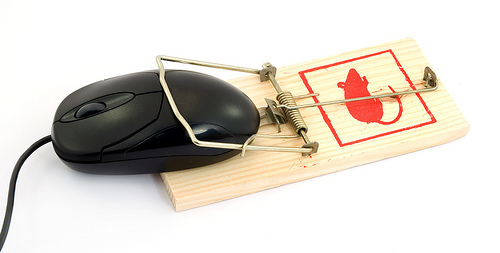}}\qquad
\subfloat[Fourth][Old label: velvet \\ \hspace*{0.14in} New label: couch]{\label{app-fig2d}\includegraphics[width = 1.3in]{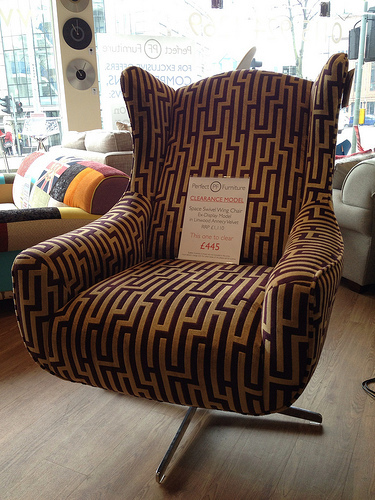}}\qquad
\subfloat[Fifth][Label: sunglass]{\label{app-fig2e}\includegraphics[width = 1.4in]{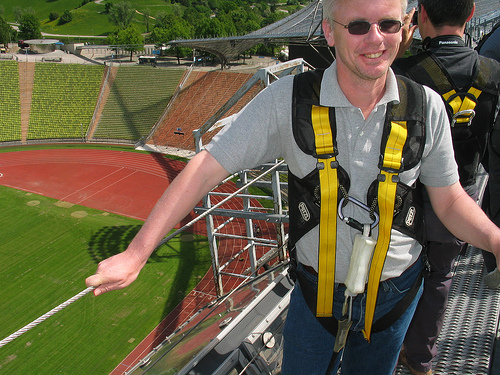}}\qquad
\subfloat[Sixth][Label: frying pan]{\label{app-fig2f}\includegraphics[width = 1.3in]{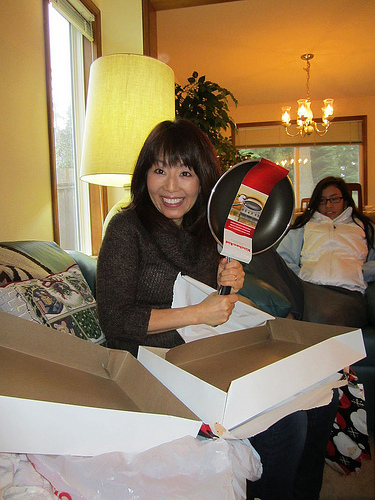}}\qquad

\caption{Example fixes for the ImageNet validation set. The first row shows label fixes by confident learning, the second row shows removed images. }
\label{app-fig2}
\end{figure*}

\cref{app-fig2} have some fixes for the ImageNet validation set. The first four examples are applied label fixes proposed by confidence learning. \cref{app-fig2a} is a correct label fix, the photo shows a \emph{racer car} that was mislabeled as \emph{sports car}. \cref{app-fig2b} is an example when the scene itself is more important than the foreground object. The original label was \emph{table} while the image was captured in a \emph{restaurant} thus the label fix was correct. \cref{app-fig2c} has two correct categories, \emph{computer mouse} and \emph{mousetrap}. The confident learning flipped these labels and since both categories are correct, the change does not cause any negative effect. \cref{app-fig2d} has an armchair that was annotated as \emph{velvet} originally and a new label \emph{couch} was assigned to it. While the couch is furniture, but the closest category to the armchairs is the \emph{chair}. The new label is not correct in this case and it should be corrected by human verification. The last two images were removed by top-5 model consensus. \cref{app-fig2e} is an example for a minuscule object and \cref{app-fig2f} has too many valid categories (\emph{frying pan, glasses, desk lamp, carton, window shade, pillow}). Both pictures provide a valid reason for removal.

\section{ImageNetV2 Validation Set Fixes}

\begin{figure*}[ph]
\centering

\subfloat[First][Old label: sports car \\ \hspace*{0.14in} New label: racer car]{\label{app-fig3a}\includegraphics[width = 1.3in]{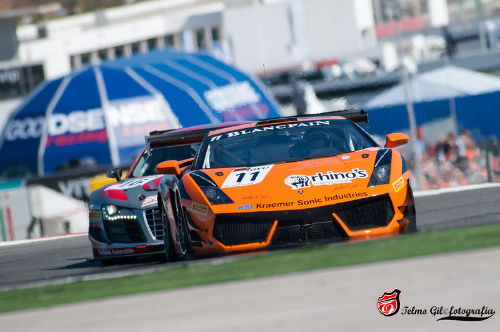}}\qquad
\subfloat[Second][Old label: table \\ \hspace*{0.14in} New label: restaurant]{\label{app-fig3b}\includegraphics[width = 1.3in]{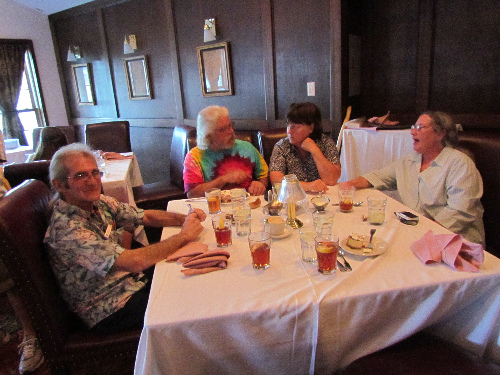}}\qquad
\subfloat[Third][Old label: bulbul \\ \hspace*{0.14in} New label: mousetrap]{\label{app-fig3c}\includegraphics[width = 1.3in]{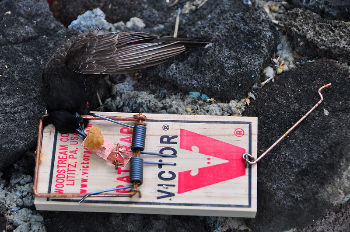}} \\
\subfloat[Fifth][Label: sunglass]{\label{app-fig3d}\includegraphics[width = 1.4in]{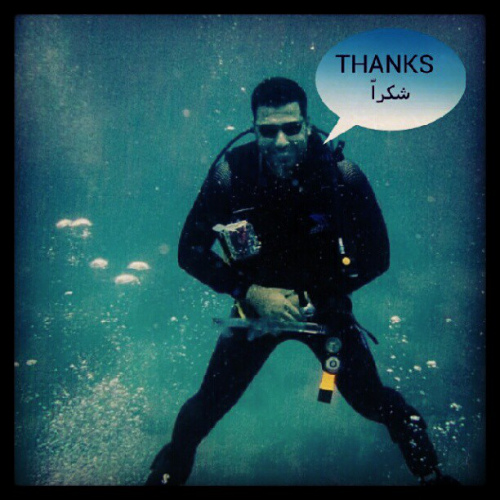}}\qquad
\subfloat[Sixth][Label: frying pan]{\label{app-fig3e}\includegraphics[width = 1.3in]{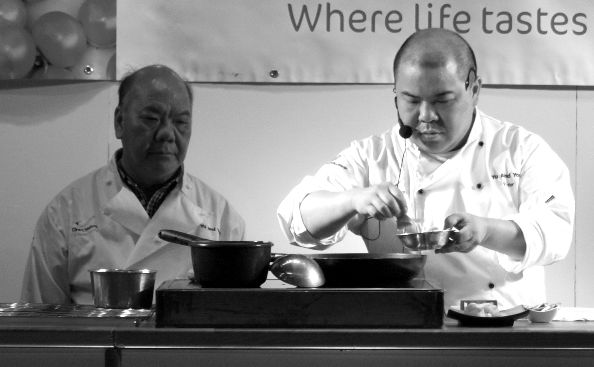}}\qquad

\caption{Example fixes for the ImageNet validation set. The first row shows label fixes by confident learning, the second row shows removed images. }
\label{app-fig3}
\end{figure*}

Despite ImageNetV2 has 10 images per categories, \cref{app-fig3} represents the same sampling problems happened for ILSVRC12 ImageNet validation set on \cref{app-fig2}. The problems were detected again by confident learning and top-5 model consensus. \cref{app-fig3a}, \cref{app-fig3b} are identical mistakes to \cref{app-fig2a} and \cref{app-fig2b}. \cref{app-fig3c} flips a valid category with \emph{mousetrap}, similar to \cref{app-fig2c}. On \cref{app-fig3d}, the sunglass is a minuscule object like on \cref{app-fig2e}. And the annotated label for \cref{app-fig3e} is \emph{frying pan} like \cref{app-fig2f}, but there are too many other categories present on the picture: tomato, plate, oven, ladle, microphone.

\newpage

\section{ILSVRC12 ImageNet Category Fixes}

\begin{table}[thp]
\caption{42 proposed categories for merging in the ILSVRC12 ImageNet dataset.}
\label{app-table1}
\vskip 0.15in
\begin{center}
\begin{adjustbox}{width=\columnwidth-1.0in,center}
\begin{sc}
\begin{tabular}{lcccr}
\toprule
Merged Category & Target Category \\
\midrule
keyboard, typewriter [n04505470] &  (computer) keyboard [n03085013] \\
bar, space bar [n04264628] & (computer) keyboard [n03085013] \\
beaker [n02815834] & (measuring) cup [n03733805] \\
maillot [n03710637] & suit, maillot [n03710721] \\
projectile, missile [n04008634] & missile [n03773504] \\
ear, spike [n13133613] & corn [n12144580] \\
(African) elephant [n02504458] & tusker [n01871265] \\
(Indian) elephant [n02504013] & tusker [n01871265] \\
sunglass [n04355933] & (sun)glasses [n04356056] \\
notebook [n03832673] & laptop [n03642806] \\
tub, vat [n04493381] & tub, bathtub [n02808440] \\
rifle, gun [n02749479] & rifle [n04090263] \\
(airplain) wing [n04592741] & (air)plane [n04552348] \\
breastplate, aegis [n02895154] & cuirass [n03146219] \\
green lizard [n01693334] & chameleon [n01682714] \\
(cassette) player [n02979186] & (tape) player [n04392985] \\
Eskimo dog, husky [n02109961] & Siberian husky [n02110185] \\
(academic) gown [n02669723] & mortarboard [n03787032] \\
(tiger) cat [n02123159] & (tabby) cat [n02123045] \\
bighorn [n02415577] & ram, tup [n02412080] \\
poodle [n02113624] & (miniature) poodle [n02113712] \\

\bottomrule
\end{tabular}
\end{sc}
\end{adjustbox}
\end{center}
\vskip -0.1in
\end{table}

\end{document}